\documentclass[10pt,twocolumn,letterpaper]{article}

\usepackage{cvpr}
\usepackage{times}
\usepackage{epsfig}
\usepackage{graphicx}
\usepackage{amsmath}
\usepackage{amssymb}
\usepackage{tabulary}
\usepackage{tabularx}
\usepackage{multirow}
\usepackage{subfig}
\usepackage{xcolor}
\usepackage[export]{adjustbox}
\usepackage[pagebackref=true,breaklinks=true,letterpaper=true,colorlinks,bookmarks=false]{hyperref}

\cvprfinalcopy % *** Uncomment this line for the final submission

\def\cvprPaperID{5842} % *** Enter the CVPR Paper ID here

\ifcvprfinal\fi

\newcolumntype{K}[1]{>{\centering\arraybackslash}p{#1}}
\DeclareMathOperator*{\argmin}{arg\,min}

\def\w{{\mathbf w}}
\def\x{{\mathbf x}}
\def\y{{\mathbf y}}

\def\P{\mathbb{P}}
\def\W{\mathbb{W}}
\def\X{{\mathbf X}}
\def\thetab{\boldsymbol\theta}

\begin{document}

%%%%%%%%% TITLE
\title{Defending against adversarial attacks by randomized diversification}

\author{Olga Taran, Shideh Rezaeifar, Taras Holotyak, Slava Voloshynovskiy\thanks{S. Voloshynovskiy is a corresponding author. The research was supported by the SNF project No. 200021\_182063.}  \\
Department of Computer Science, University of Geneva \\
7, route de Drize, 1227 Carouge, Switzerland\\
{\tt\small \{olga.taran, shideh.rezaeifar, taras.holotyak, svolos\}@unige.ch}
%\and
}

\maketitle
\thispagestyle{empty}

%%%%%%%%% ABSTRACT
\begin{abstract}
The vulnerability of machine learning systems to adversarial attacks questions their usage in many applications. In this paper, we propose a randomized diversification as a defense strategy. We introduce a multi-channel architecture in a \textit{gray-box} scenario, which assumes that the architecture of the classifier and the training data set are known to the attacker. The attacker does not only have access to a secret key and to the internal states of the system at the test time. The defender processes an input in multiple channels. Each channel introduces its own randomization in a special transform domain based on a secret key shared between the training and testing stages. Such a transform based randomization with a shared key preserves the gradients in key-defined sub-spaces for the defender but it prevents gradient back propagation and the creation of various bypass systems for the attacker. An additional benefit of multi-channel randomization is the aggregation that fuses soft-outputs from all channels, thus increasing the reliability of the final score. The sharing of a secret key creates an information advantage to the defender. Experimental evaluation demonstrates an increased robustness of the proposed method to a number of known state-of-the-art attacks.

\end{abstract}

%%%%%%%%% BODY TEXT
\section{Introduction}
Besides remarkable and impressive achievements, many machine learning  systems are vulnerable to adversarial attacks \cite{goodfellow6572explaining}. The adversarial attacks attempt at tricking a decision of a classifier by introducing bounded and invisible perturbations to a chosen target image. This weakness seriously questions the usage of the machine learning in many security- and trust-sensitive domains.

Many researchers have proposed various defense strategies and countermeasures to defeat adversarial attacks.  However, the growing number of defenses naturally stimulates the invention of new and even more universal attacks.  An overview and classification of the most efficient attacks and defenses are given in \cite{yuan2017adversarial, taran2018bridging}.

\begin{figure}[t!]
\centering
\includegraphics[width=1\linewidth]{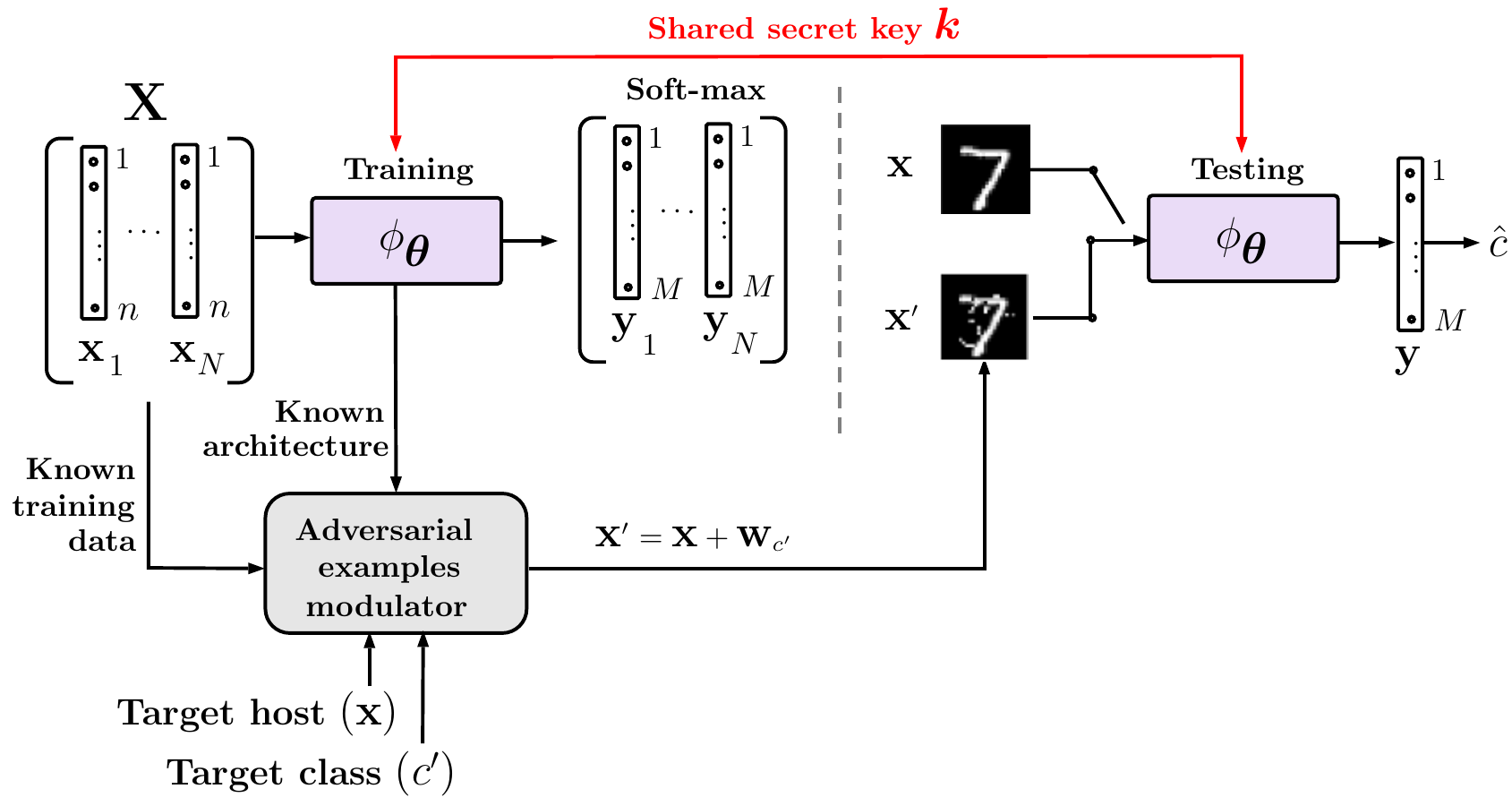}
\caption{Setup under investigation: the attacker knows the labeled training data set $\X$ and the system architecture but he does not have access to secret key $k$ of the defender shared between the training and testing. }
\label{fig:setup under investigation}
\end{figure}

%Despite the fact that, in general, the defender has the final word, there exist several important open issues that make the overwhelming majority of defense mechanisms not enough reliable. 
In this paper, we consider a "game" between the defender and the attacker according to the diagram presented in Figure \ref{fig:setup under investigation}. 
\begin{figure*}[t!]
\centering
\includegraphics[width=0.65\linewidth]{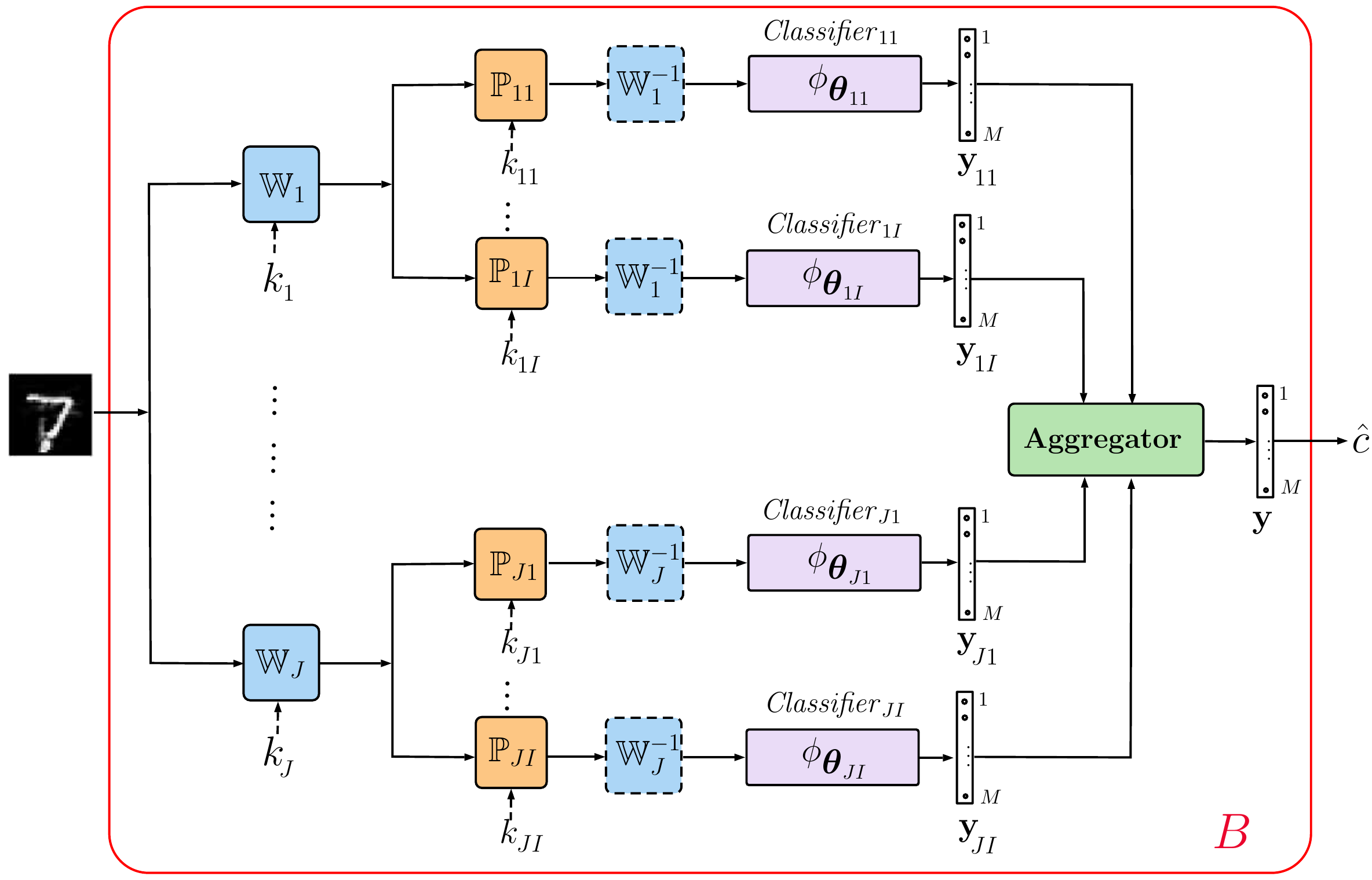}
\caption{Generalized diagram of the proposed multi-channel classifier.}
\label{fig:principal classification scheme}
\end{figure*}

The defender has access to the classifier $\phi_{\thetab}$ and the training data set $\X$. The defender shares a secret key $k$ between training and testing. The classifier outputs a soft-max vector $\y$ of length $M$, where $M$ corresponds to the total number of classes, and each $y_c$, $1 \le c \le M$ is treated as a probability that a given input $\x$ belongs to a class $c$. The trained classifier $\phi_{\thetab}$ is used during testing.

The attacker in the \textit{white-box} scenario has full knowledge about the classifier architecture, defense mechanisms, training data and, quite often, can access the trained parameters of the classifier. In the \textit{gray-box} scenario, considered in this paper, the attacker knows the architecture of the classifier, the general defense mechanism and has access to the same training data $\X$ \cite{chen2017zoo, yuan2017adversarial}. Using the above available knowledge, the attacker can generate a \textit{non-targeted} or \textit{targeted}, with respect to a specified class $c^\prime$, adversarial perturbation $\w_{c^\prime}$. The attacker produces an adversarial example by adding this perturbation to the target host sample $\x$ as $\x^\prime = \x + \w_{c^\prime}$. The adversarial example is presented to the classifier at test time in an attempt to trick the classifier $\phi_{\thetab}$ decision.

Without pretending to be exhaustive in our overview, we group existing defense strategies into three major groups:
\begin{enumerate}
    \item {\bf Non key-based defenses}: This group includes the majority of state-of-the-art defense mechanisms based on detection and rejection, adversarial retraining, filtering and regeneration, etc. \cite{taran2018bridging}. Besides the broad diversity of these methods, a common feature and the main disadvantage of these approaches is an absence of "cryptographic" elements, like for example a secret key, that would allow to create an information advantage of the defender over the attacker.
    
    \item {\bf Defense via randomization and obfuscation}: The defense mechanisms of this group are mainly based on the ideas of randomization avoiding the reproducible and repeatable use of parameters of the trained system. This includes gradient masking \cite{athalye2018obfuscated} and introducing an ambiguity via different types of key-free randomization. The example of such randomization can be noise addition at different levels of the system \cite{you2018adversarial}, injection of different types of randomization like, for example, random image resizing or padding \cite{xie2017mitigating} or randomized lossy compression \cite{das2018shield}, etc.
   
    The main disadvantage of this group of defense strategies consists in the fact that the attacker can bypass the defense blocks or take this ambiguity into account during the generation of the adversarial perturbations \cite{athalye2018obfuscated}. Additionally, the classification accuracy is degraded since the classifier is only trained on average for different sets of randomization parameters unless special ensembling or aggregation is properly applied to compensate this loss. However, even in this case the mismatch between the training and testing stages can only ensure the performance on average whereas one is interested to have the guaranteed performance for each realization of randomized parameters. Unfortunately, this is not achievable without the common secret sharing between the training and testing.
    
    \item {\bf Key-based defenses}:  The third group generalizes the defense mechanisms, which include a randomization explicitly based on a secret key that is shared between training and testing stages. For example, one can mention the use of random projections \cite{vinh2016training}, the random feature sampling \cite{chen2018secure} and the key-based transformation \cite{taran2018bridging}, etc. 
    
    Nevertheless, the main disadvantage of the known methods in this group consists of the loss of performance due to the reduction of useful data that should be compensated by a proper diversification and corresponding aggregation.
    \end{enumerate}

In this paper, we target further extension of the key-based defense strategies based on the cryptographic principles to create an information advantage of the defender over the attacker yet maximally preserving the information in the classification system. The generalized diagram of the proposed system is shown in Figure \ref{fig:principal classification scheme}. It has two levels of randomization, each of which can be based on unique secret keys. An additional robustification is achieved via the aggregation of the soft outputs of the multi-channel classifiers trained for their own randomizations. As it will be shown throughout the paper, usage of multi-channel architecture diminishes the efficiency of attacks. 

The main contribution of this paper is twofold: 
\vspace{-\topsep}
\begin{itemize}
  \setlength{\parskip}{0pt}
  \setlength{\itemsep}{0pt plus 1pt}
\item A new multi-channel classification architecture with defense strategy against \textit{gray-box} attacks based on the cryptographic principle.
\item An investigation of the efficiency of the proposed approach on three standard data sets for several classes of well-known adversarial attacks.
\end{itemize}

The remainder of this paper is organized as follows: Section \ref{sec:Multi-channel classification algorithm} introduces a new multi-channel classification architecture. Section \ref{sec:classification via several permutations} provides an extension of the defense strategy based on the data independent permutation proposed in \cite{taran2018bridging} to multi-channel architecture. The efficient key-based data independent transformation is investigated in Section \ref{sec:DCT sign permutation}. The filtering by a hard-thresholding in the secret domain is analyzed in Section \ref{sec:zero filling as a defense strategy}. Section \ref{sec:conclusions} concludes the paper.

%-------------------------------------------------------------------------
\section{Multi-channel classification algorithm}
\label{sec:Multi-channel classification algorithm}

A multi-channel classifier, which forms the core of the proposed architecture, is shown in Figure \ref{fig:principal classification scheme}. It consists of four main building blocks:
\begin{enumerate}
\item Pre-processing of the input data in a \textit{transform domain} via a mapping $\W_{j}$, $1 \le j \le J$. In general, the  transform $\W_{j}$ can be any linear mapper. For example it can be a random projection or belong to the family of orthonormal transformations ($\W_{j}\W^{T}_{j} = \mathbb{I}$) like DFT (discrete Fourier transform), DCT (discrete cosines transform), DWT (discrete wavelet transform), etc. Moreover, $\W_{j}$ can also be a learnable  transform. However, it should be pointed out that from the point of view of the robustness to adversarial attacks, the data independent transform $\W_{j}$ is of interest to avoid key-leakage from the training data. Furthermore, $\W_{j}$ can be based on a secret key $k_j$.
 
\item \textit{Data independent processing} $\P_{ji}$, $1 \le i \le I$ presents the second level of randomization and serves as a defense against gradient back propagation to the direct domain.

One can envision several cases. As shown in Figure \ref{fig:p_sampling}, $\P_{ji} \in \{0, 1\}^{l \times n}$, $l < n$, presents a lossy sampling of the input signal of length $n$, as considered in \cite{chen2018secure}. In  Figure \ref{fig:p_permutation}, $\P_{ji} \in \{0, 1\}^{n \times n}$ is a lossless permutation, similar to \cite{taran2018bridging}. Finally, in Figure \ref{fig:p_sign_permutation}, $\P_{ji} \in \{-1, 0, +1\}^{n \times n}$ corresponds to sub-block sign flipping. The yellow color highlights the key defined region of key-based sign flipping. This operation is reversible and thus lossless for an authorized party. Moreover, to make the \textit{data independent processing} irreversible for the attacker, it is preferable to use a $\P_{ji}$ based on secret key $k_{ji}$.

\begin{figure}[t!]
     \centering
     \subfloat[][randomized sampling]{\quad\includegraphics[width=0.325\linewidth,valign=t]{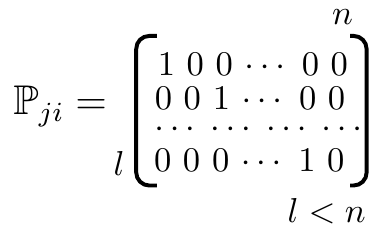}\label{fig:p_sampling}\quad} 
     \subfloat[][randomized permutation]{\quad\includegraphics[width=0.325\linewidth,valign=t]{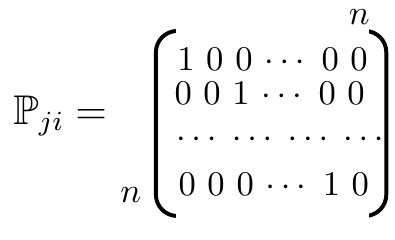}\label{fig:p_permutation}\quad}  
    
     \subfloat[][randomized sign flipping in the sub-block defined in orange ]{\quad\includegraphics[width=0.4\linewidth,valign=t]{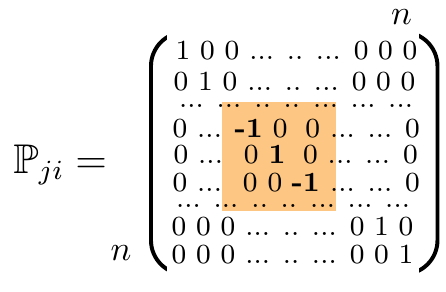}\label{fig:p_sign_permutation}\quad}   
     \caption{Randomized transformation $\P_{ji}$, $1 \le j \le J, \; 1 \le i \le I$ examples. All transforms are key-based.}
     \label{fig:dct_sign_permut}
\end{figure}

\item \textit{Classification block} can be represented by any family of classifiers. However, if the classifier is designed for classification of data in the direct domain then it is preferable that it is preceded by $\W_{j}^{-1}$.

\item \textit{Aggregation block} can be represented by any operation ranging from a simple summation to learnable operators adapted to the data or to a particular adversarial attack.
\end{enumerate}

As it can be seen from Figure \ref{fig:principal classification scheme}, the chain of the first 3 blocks can be organized in a parallel multi-channel structure that is followed by one or several \textit{aggregation blocks}. The final decision about the class is made based on the aggregated result. The rejection option can be also naturally envisioned.

The training of the described algorithm can be represented as:
\begin{equation}
\label{eq:general_formula1}
(\boldsymbol{\hat{\vartheta}}, \{\boldsymbol{\hat{\theta}_{ji}} \}) = \argmin_{\boldsymbol{\vartheta}, \{\boldsymbol{\theta}_{ji}\}} \sum_{t=1}^T \sum_{j=1}^J \sum_{i=1}^{I_j} \mathcal{L}(\y_t, A_{\boldsymbol{\vartheta}}(\phi_{{\boldsymbol\theta}_{ji}}(f(\x_t)))), \\
\end{equation}
with: 
\begin{displaymath}
\label{eq:general_formula2}
f(\x_t) = \W_j^{-1}\P_{ji}\W_j\x_t,
\end{displaymath}
where $\mathcal{L}$ is a classification loss, $\y_t$ is a vectorized class label of the sample $\x_t$, $A_{\boldsymbol{\vartheta}}$ corresponds to the aggregation operator with parameters ${\boldsymbol{\vartheta}}$, $\phi_{\thetab_{ji}}$ is the $i$th classifier of the $j$th channel, $\thetab$ denotes the parameters of the classifier, $T$ equals to the number of training samples, $J$ is the total number of channels and $I_j$ equals to the number of classifiers per channel that we will keep fixed and equals to $I$.

The attacker might discover the secret keys $k_j$ and/or $k_{ji}$ or make the full system end-to-end differentiable using the Backward Pass Differentiable Approximation technique proposed in \cite{athalye2018obfuscated} or via replacing the key-based blocks by the bypass mappers. To avoid such a possibility, we restrict the access of the attacker to the internal results within the block $B$. This assumption corresponds to our definition of the gray-box setup.

In the proposed system, we will consider several practical simplifications leading to information and complexity advantages for the defender over the attacker: 
\vspace{-\topsep}
\begin{itemize}
  \setlength{\parskip}{0pt}
  \setlength{\itemsep}{0pt plus 1pt}
\item The defender training can be performed per channel independently until the \textit{aggregation block}. At the same, the attacker should train and back propagate the gradients in all channels simultaneously or at least to guarantee the majority of wrong scores after aggregation.
\item The blocks of \textit{data independent processing} $\P_{ji}$ aim at preventing gradient back propagation into the direct domain but the classifier training is adapted to a particular $\P_{ji}$ in each channel. 
\item It will be shown further by the numerical results that the usage of the multi-channel architecture with the following aggregation stabilizes the results' deviation due to the use of randomizing or lossy transformations $\P_{ji}$, if such are used.
\item The right choice of the \textit{aggregation} operator $A_{\boldsymbol{\vartheta}}$ provides an additional degree of freedom and increases the security of the system through the possibility to adapt to specific types of attacks.
\item Moreover, the overall security level considerably increases due to the independent randomization in each channel. The main advantage of the multi-channel system consists in the fact that each channel can have an adjustable amount of randomness, that allows to obtain the required level of defense against the attacks. In a one-channel system the amount of randomness can be either insufficient to prevent the attacks or too high which leads to classification accuracy loss. Therefore, having a channel-wise distributed randomness is more flexible and efficient for the above trade-off. 
\end{itemize}

The described generalized multi-channel architecture provides a variety of choices for the transform operators $\W$ and data independent processing $\P_{ji}$. In Section \ref{sec:classification via several permutations}, we will consider a variant with multiple $\P_{ji}$ in the form of the considered permutation in the direct domain $\W_j=\mathbb{I}$. In Section \ref{sec:DCT sign permutation}, we will investigate a sign flipping operator $\P_{ji}$ for the common DCT operator $\W$. Section \ref{sec:zero filling as a defense strategy} will be dedicated to the investigation of a denoising  version of $\P_{ji}$ based on hard-thresholding in a secret sub-space of the DCT domain $\W_j$.

%-------------------------------------------------------------------------
\section{Classification with multi-channel permutations in the direct domain }
\label{sec:classification via several permutations}

The simplest case of randomized diversification can be constructed for the direct domain with the permutation of input pixels. In fact, the algorithm proposed in  \cite{taran2018bridging} reflects this idea for a single channel. However, despite the reported efficiency of the proposed defense strategy, a single channel architecture is subject to a drop in classification accuracy, even for the original, i.e., non-adversarial, data. 

%Such a behaviour is  explained by the gradient masking. 

Therefore, this paper investigates the performance of a permutation-based defense in a multi-channel setting.

\subsection{Problem formulation}
\begin{figure}[t!]
\centering
\includegraphics[width=0.8\linewidth]{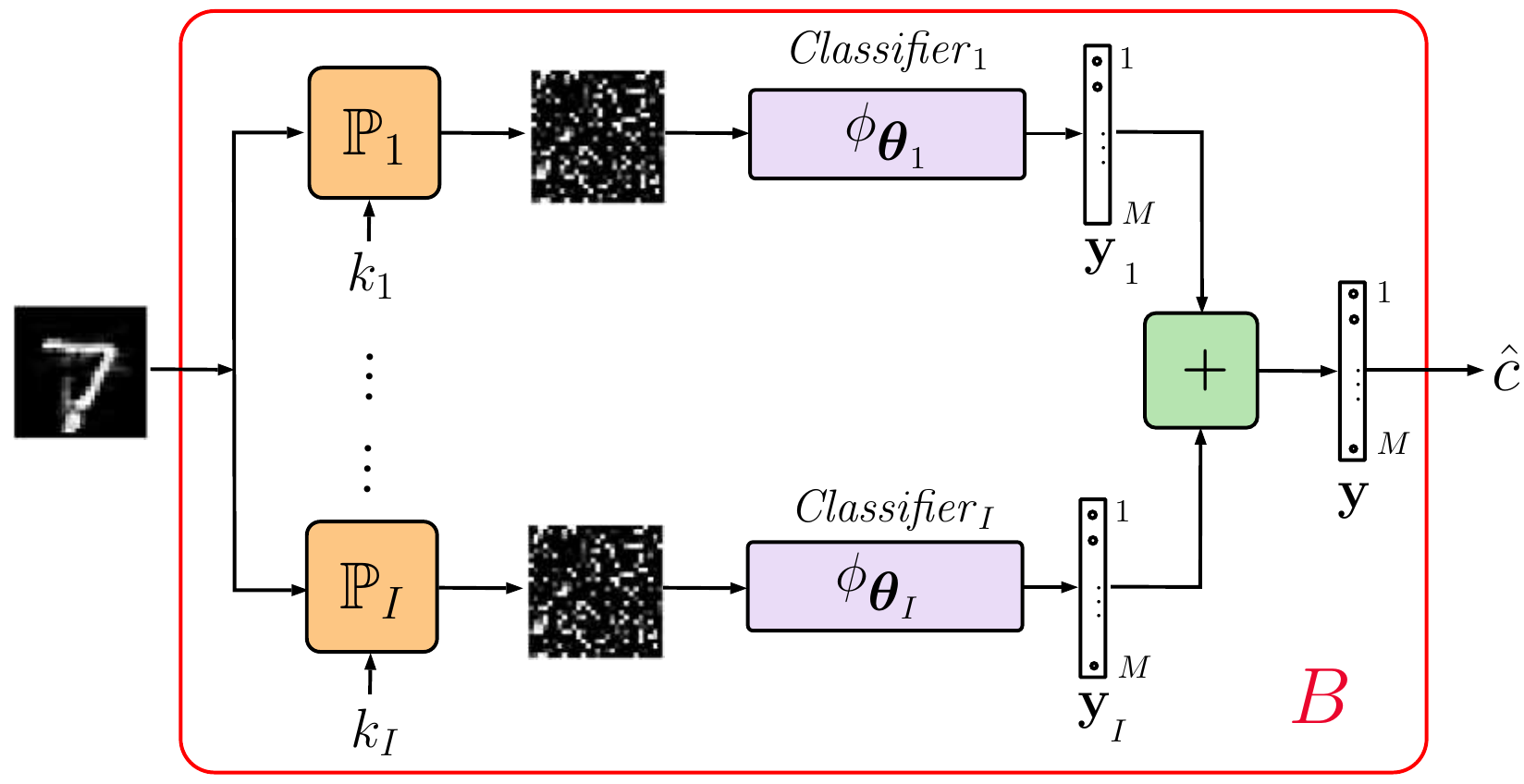}
\caption{Classification via multi-channel permutations in the direct domain.}
\label{fig:classification_via_several_permut}
\end{figure}

The generalized diagram of the corresponding extended multi-channel approach is illustrated in Figure \ref{fig:classification_via_several_permut}. The permutation in the direct domain implies that $\W_j = \mathbb{I}$ with $J=1$ and $I$ permutation channels. Therefore, each channel $1 \leq i \leq I$ has only one data independent permutation block $\P_{i}$ represented by a lossless permutation of the input signal $\x \in \mathbb{R}^{n \times n\times m}$ in the direct domain, where  $n$ corresponds to the size of the input image and  $m$ is the number of the channels (colors) in this image. Thus, the permutation matrix $\P_{i}$ is a matrix of size $n \times n$, generated from a secret key $k_i$, whose entries are all zeros except for a single element of each row, which is equal to one. In addition, as illustrated in Figure \ref{fig:p_permutation}, all non-zero entries are located in different columns. For our experiments, we assume that $\P_{i}$ is the same for each input image color channel but it can be a different one in the general case to increase the security of the system. As an aggregation operator $A$, we use a summation for the sake of simplicity and interpretability. The aggregated result represents a $M$-dimensional vector $\y$ of real non-negative values, where $M$ equals to the number of classes and each entry $y_{c}$ is treated as a probability that a given input $\x$ belongs to the class $c$.

Under the above assumptions, the optimization problem (\ref{eq:general_formula2}) reduces to \footnote{https://github.com/taranO/defending-adversarial-attacks-by-RD}: 
\begin{equation}
\{\boldsymbol{\hat{\theta}}_i\} = \argmin_{\{{\boldsymbol\theta}_i\}} \sum_{t=1}^T \sum_{i=1}^I \mathcal{L}(\y_t, A(\phi_{\boldsymbol{\theta}_{i}}(\P_{i}\x_t))).
\label{eq:several_permut_formula}
\end{equation}
%

%It should be noted that each classifier $\phi_{\boldsymbol{\theta}_{j}}$ is trained independently to minimize the channel $j$ loss:
%%
%\begin{equation}
%\boldsymbol{\hat{\theta}}_j = \argmin_{{\boldsymbol\theta}_j} \sum_{t=1}^T \mathcal{L}(c_t, %\phi_{\boldsymbol{\theta}_{j}}(\P_{j}\x_t))
%\label{eq:several_permut_formula}
%\end{equation}
%% 

\subsection{Numerical results}
\label{subsec_several_permut_res}
%
% ---------------------------------------
\begin{table}[t!]
{\footnotesize
\renewcommand*{\arraystretch}{1.25}
\begin{tabular}{K{1.5cm}cccccc} \hline
\multirow{ 2}{*}{Data type}  & \multicolumn{6}{c}{$I$} \\ \cline{2-7}
& 1 & 5 & 10 & 15 & 20 & 25 \\ \hline
\multicolumn{7}{c}{\textbf{MNIST}: original classifier error is 1\%} \\ \hline
Original                  & 2.83 & 1.73 & 1.37 & 1.43 & 1.57 & 1.4 \\ 
\textit{CW $\ell_2$}      & 8.85 & 4.56 & 3.82 & 3.53 & 3.55 & 3.51 \\
\textit{CW $\ell_0$}      & 13.87 & 5.98 & 4.98 & 4.69 & 4.47 & 4.4 \\ 
\textit{CW $\ell_\infty$} & 11.67 & 4.72 & 4.03 & 3.87 & 3.59 & 3.69 \\ \hline
\multicolumn{7}{c}{\textbf{Fasion-MNIST}: original classifier error is 7.5\%} \\ \hline
Original                  & 11.40 & 9.4   & 9.27 & 9.2 & 9.23  & 9.2 \\ 
\textit{CW $\ell_2$ }     & 12.16 & 10.15 & 9.78 & 9.41 & 9.49 & 9.4 \\  
\textit{CW $\ell_0$}      & 13.45 & 10.15 & 9.62 & 9.56 & 9.82 & 9.63 \\  
\textit{CW $\ell_\infty$} & 11.99 & 9.72  & 9.69 & 9.24 & 9.26 & 9.32 \\ \hline
\multicolumn{7}{c}{\textbf{CIFAR}: original classifier error is 21\%} \\ \hline
Original                  & 47.03 & 41.47 & 40.2 & 39.8 & 39.2 & 39       \\ 
\textit{CW $\ell_2$ }     & 47.76 & 41.82 & 39.83 & 39.59 & 39.4 & 39.04  \\ 
\textit{CW $\ell_0$}      & 48.39 & 42.27 & 40.87 & 39.73 & 39.85 & 39.76 \\  
\textit{CW $\ell_\infty$} & 47.41 & 42.12 & 40.53 & 39.58 & 39.62 & 39.21 \\ \hline
\end{tabular}
}
\caption{Classification error (\%) on the first 1000 test samples for $I$-channel system  with the direct domain permutation.}
\label{tab:mnist_classification_error}
\end{table}

To reveal the impact of multi-channel processing, we compare our results with an identical single channel system reported in \cite{taran2018bridging}. For each classifier  $\phi_{\boldsymbol{\theta}_{i}}$ we use exactly the same architecture as mentioned in Table 2 in \cite{taran2018bridging}\footnote{The  Python code for generating adversarial examples is available at https://github.com/carlini/nn\_robust\_ attacks}. Moreover, taking into account that the generation of adversarial examples is quite a slow process, as well as in \cite{taran2018bridging}, we verify our approach on the first 1000 test samples of the MNIST \cite{lecun2010mnist} and Fashion-MNIST \cite{xiao2017fashion} data sets. Additionally, we investigate the CIFAR-10 data set \cite{krizhevsky2014cifar}. 

The obtained results are given in Table \ref{tab:mnist_classification_error}. For all data sets, a single channel set up with $I=1$ corresponds to the results of the approach proposed in \cite{taran2018bridging} and \textit{CW} denotes the attacks proposed by Carlity and Wagner in \cite{carlini2017towards}.

As one can note from Table \ref{tab:mnist_classification_error}, increasing the number of channels leads to a  decrease of the classification error. In the case of the MNIST data set, our multi-channel algorithm allows to reduce the error on the original non-attacked data in 2 times, from 2.8\% to 1.4\%. For the attacked data, the classification error decreases 2.5 times from almost 9-14\% to 3.5-4.5\%. In case of the Fashion-MNIST data set, one can observe a similar dynamic, namely, the classification error decreases from 11.5-13.5\% to 9-9.5. For the CIFAR-10 data set using the multi-channel architecture allows to reduce the error from 47-48\% to only about 39.5\%. The CIFAR-10 natural images are more complex in comparison to the MNIST and Fasion-MNIST and the introduced permutation destroys local correlations. This has a direct impact on classifier performance.

%In general, the obtained results show that the proposed classification with  multi-channel permutations in the direct domain improves the performance and to stabilize the classification accuracy.% on the input signals protected against the adversarial attacks via the data-independent transformations that lead to the gradient masking and violation. Moreover, using of the several secret keys makes the system more robust to the attack aims at guessing the key. %Additionally, it should be emphasized that the using of multi-channel architecture allows to stabilize the classification accuracy. %Moreover, despite the big number of used classifiers the test procedure is quite fast. 

%
\begin{figure}[t!]
     \centering
     \subfloat[][coordinate domain]{\includegraphics[width=0.37\linewidth]{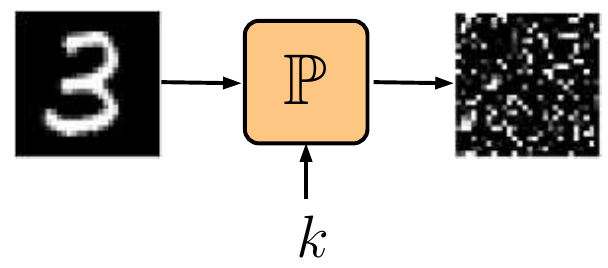}\label{fig:permut_coordinate}}
     \hspace{0.3cm}
     \subfloat[][DCT domain]{\includegraphics[width=0.55\linewidth]{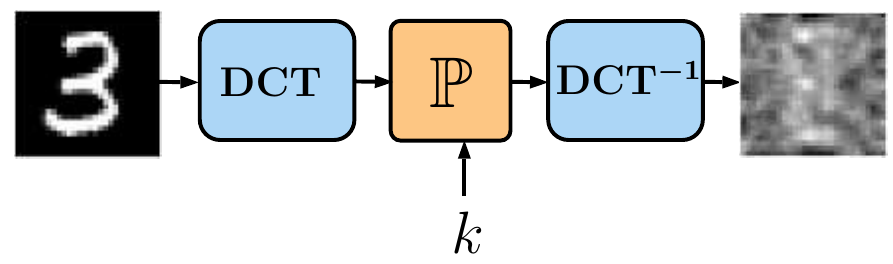}\label{fig:permut_dct}}
     \caption{Global permutation in the coordinate domain (a) and DCT based encoding using key-based sign flipping (b).}
     \label{fig:global_permuts}
\end{figure}
%
%-------------------------------------------------------------------------
\section{Classification with multi-channel sign permutation in the DCT domain}
\label{sec:DCT sign permutation}

The results obtained in Section \ref{sec:classification via several permutations} for the CIFAR-10 data set show a high sensitivity to the gradient perturbations that degrade the performance of the classifier. In this Section we investigate the other data independent processing functions $\P_{ji}$ based on a secret key $k_{ji}$ preserving the gradient in a special way that is more suitable for the classification of complex natural images. We will consider a general scheme for demonstrative purposes to justify that the permutations should be localized rather than global.

\subsection{Global permutation}

We will consider sign flipping in the DCT domain as a basis for the multi-channel randomization. 
For the visual comparison of the effect of global permutation in the coordinate domain versus global sign flipping in the DCT domain, we show an example in Figure \ref{fig:global_permuts}. From this Figure one can note that the permutation in the coordinate domain  disturbs the local correlation in the image that will impact the local gradients. In turn, this might impact the training of modern  classifiers that are mostly based on gradient techniques. At the same time, preservation of the gradients makes the data more vulnerable to adversarial attacks. Keeping this in mind, we can conclude that the global DCT sign permutation also "randomizes" the images but, in contrast to the permutation in direct domain, it keeps the local correlation. 

To answer the question whether the preservation of local correlation at the randomization can help preserve the loss of the gradients, we investigate the global DCT sign permutation for the classification architecture shown in Figure \ref{fig:classification_via_several_permut} with $\W_j$  is DCT and $\P_{i} \in \{-1,1\}^{n \times n}$. It should be noted that the transform $\W_j$ is fixed for all channels. Therefore, the secrecy part consists in the key-based flipping of DCT coefficients' signs.

In the experimental results we obtained the classification accuracy to be  very close to the results represented in Table \ref{tab:mnist_classification_error}. For the sake of space, we do not present this table in the paper. Nevertheless, we can conclude that the global sign permutation in the DCT domain does not improve the previous situation with the global permutation in direct domain.

\begin{figure}[t!]
     \centering
     \subfloat[][sub-bands]{\quad\includegraphics[width=0.12\linewidth,valign=b]{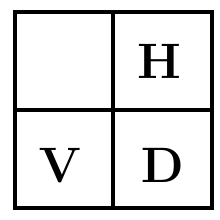}\label{fig:dct_subbands}\quad}
     \subfloat[][original]{\quad\includegraphics[width=0.125\linewidth,valign=b]{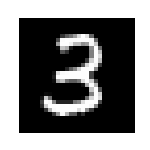}\label{fig:permut_dct_original}\quad}     

     \subfloat[][V]{\quad\includegraphics[width=0.125\linewidth]{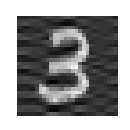}\label{fig:permut_dct_v}\quad}
     \subfloat[][H]{\quad\includegraphics[width=0.125\linewidth]{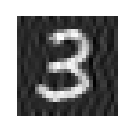}\label{fig:permut_dct_h}\quad}     
     \subfloat[][D]{\quad\includegraphics[width=0.125\linewidth]{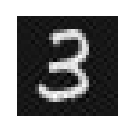}\label{fig:permut_dct_d}\quad}     
     \subfloat[][V+H+D]{\quad\includegraphics[width=0.125\linewidth]{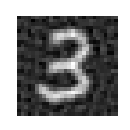}\label{fig:permut_dct_dhv}\quad}   
     \caption{Local randomization in the DCT sub-bands by key-based sign flipping.}
     \label{fig:dct_sign_permut}
\end{figure}
%
% --------------
\subsection{Local permutation}

Taking into account the above observation, we investigate the behaviour of the local DCT sign permutations, i.e., we will use a global DCT transform but will flip the signs only for the selected number of coefficients as shown in Figure \ref{fig:permut_dct_v}.

The general idea, illustrated in Figure \ref{fig:dct_sign_permut}, consists in the fact that the DCT domain can be split into overlapping or non-overlapping sub-bands of different size. In our case, for the simplicity and interpretability, we split the DCT domain into 4 sub-bands, namely, (1) top left that represents the low frequencies of the image, (2) vertical, (3) horizontal and (4) diagonal sub-bands. After that we apply the DCT sign flipping as randomization in each sub-band keeping all other sub-bands unchanged and apply the inverse DCT transform. The corresponding illustrative examples are shown in Figures \ref{fig:permut_dct_v} - \ref{fig:permut_dct_d}. Finally, we apply the DCT sign permutation in 3 sub-bands. The  corresponding result is shown in Figure \ref{fig:permut_dct_dhv}. It is easy to see that local DCT sign flipping applied in one individual sub-band creates a specific oriented distortion due to the specificity of chosen sub-bands but preserves the local image content quite well. The simultaneous permutation of 3 sub-bands creates more degradation which might be undesirable and can have a negative influence on the classification accuracy. 

\begin{figure}[t!]
\centering
\includegraphics[width=1\linewidth]{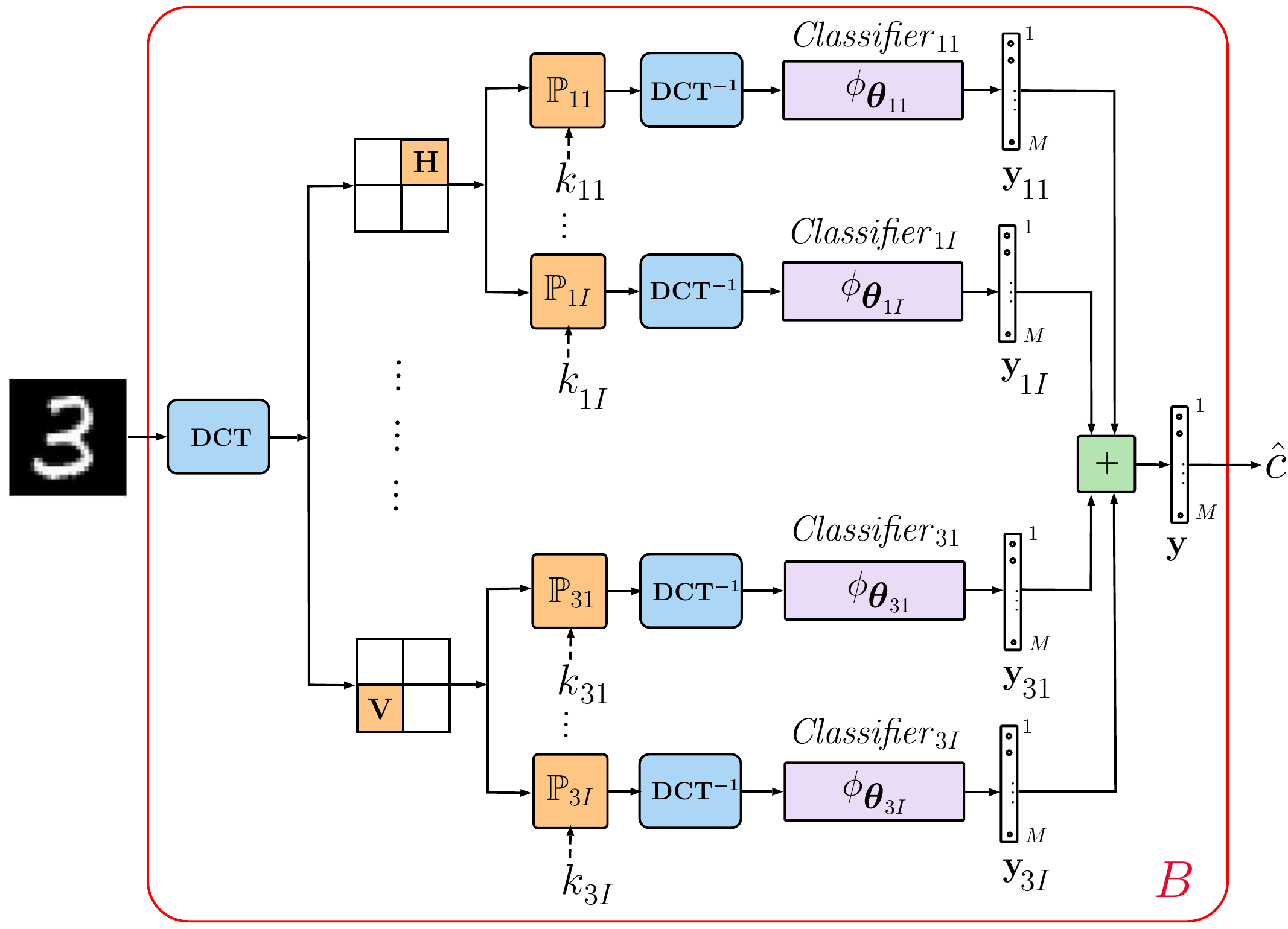}
\caption{Classification with local DCT sign permutations.}
\label{fig:classification_via_dct_permut}
\end{figure}

To investigate the behaviour of the local DCT sign permutations we use the multi-channel architecture shown in Figure \ref{fig:classification_via_dct_permut}. It is a three-channel model with $I$ sub-channels. As a $\W_{j}$ we use a standard DCT transform. The sub-channels' data independent processing blocks $\P_{ji} \in \{-1, 0, 1\}^{n \times n}$ are based on the individual secret keys $k_{ji}$ and are represented by the matrices that allow to change the elements' signs only in the sub-band of interest, like illustrated in Figure \ref{fig:p_sign_permutation}. In general case, the sub-bands can be overlapping or non-overlapping and have different positions and sizes. As discussed, we use only 3 non-overlapping sub-bands of equal size as illustrated in Figure \ref{fig:dct_subbands}. The architecture of the classifiers $\phi_{\boldsymbol{\theta}_{ji}}$ is identical to the ones used in Section \ref{sec:Multi-channel classification algorithm}. As an aggregation operator we use a simple summation.

The corresponding optimization problem becomes: 
\begin{equation}
\{\boldsymbol{\hat{\theta}}_{ji}\} = \argmin_{\{\boldsymbol{\theta}_{ji}\}} \sum_{t=1}^T \sum_{j=1}^3 \sum_{i=1}^I \mathcal{L}(\y_t, A(\phi_{{\boldsymbol\theta}_{ji}}(\P_{ji}\x_t))).
\label{eq:local_dct_formula}
\end{equation}
%
%-------------------------------------------------------------------------
\subsection{Numerical results}
%

%To investigate the influence of increasing of the number of channels we use several key-transformation in each sub-band. 
The results obtained for the architecture proposed in Figure \ref{fig:classification_via_dct_permut} are shown in Table \ref{tab:dct_sign_classification_error}. The column "Classical" corresponds to the results of the one-channel classical classifier for the original non-permuted data, that is referred to as the classical scenario.

It should be pointed out that in the previous experiments we observed a drop in the classification accuracy even for the original non-attacked data. In the proposed scheme with the 12 and 15 sub-channels, the obtained classification error on the adversarial examples corresponds to those of the original data and, in some cases, is ever lower. For example, we obtained a 2 times decrease in the classification error on the MNIST for the original data in comparison to the classical architecture.

The CIFAR-10 data set presents a particular interest for us as a data set with natural images. For \textit{CW $\ell_2$} and \textit{CW $\ell_\infty$} attacks the classification error is the same as in the case of the classical scenario on the original data. This demonstrates that the proposed method does not cause a degradation in performance due to the introduced defense mechanism. In case of \textit{CW $\ell_0$} attack there is only about 2\% of successful attacks. 
%Taking into account that the saturation is not observed and further increasing of the number of classifiers (via increasing either the number of channels or via increasing the number of sub-channel permutations) might lead to even better results.

In the case of the Fashion-MNIST data set, the obtained results are better than the results for the permutation in the direct domain given in Table \ref{tab:mnist_classification_error}. For the original non-attacked data the classical scenario accuracy is achieved. For the attacked data the classification error exceeds the level of those on the original data only with 1-2\%. 

% ---------------------------------------
\begin{table}[t!]
{\footnotesize
\renewcommand*{\arraystretch}{1.25}
\centering
\begin{tabular}{ccccccc} \hline
\multirow{ 2}{*}{Data type}  & \multirow{ 2}{*}{Classical} & \multicolumn{4}{c}{$J \cdot I$} \\ \cline{3-7}
& & 3 & 6 & 9 & 12 & 15 \\ \hline
\multicolumn{6}{c}{\textit{MNIST}}  \\ \hline
Original                  & 1     & 0.5  & 0.5   & 0.5  & 0.5  & 0.5   \\ 
\textit{CW $\ell_2$ }     & 100   & 6.28 & 5.34  & 4.66 & 4.44 & 4.73  \\ 
\textit{CW $\ell_0$}      & 100   & 19.3 & 18.48 & 17.6 & 16.7 & 17.42 \\ 
\textit{CW $\ell_\infty$} & 99.99 & 2.81 & 2.37  & 2.22 & 2.12 & 2.06  \\  \hline
\multicolumn{6}{c}{\textit{Fashion-MNIST}}  \\ \hline
Original                  & 7.5  & 8.1   & 7.4  & 7.6   & 7.2  & 7.4 \\ 
\textit{CW $\ell_2$ }     & 100  & 9.27  & 8.67 & 8.87  & 8.62 & 8.62 \\ 
\textit{CW $\ell_0$}      & 100  & 10.62 & 9.99 & 10.13 & 9.87 & 9.86 \\ 
\textit{CW $\ell_\infty$} & 99.9 & 9.2   & 8.41 & 8.66  & 8.47 & 8.49 \\  \hline
\multicolumn{6}{c}{\textit{CIFAR-10}}  \\ \hline
Original                  & 21  & 21.2  & 19.6  & 19.5  & 18.6  & 19.2 \\ 
\textit{CW $\ell_2$}      & 100 & 22.42 & 21.3  & 21.04 & 20.79 & 20.92 \\ 
\textit{CW $\ell_0$}      & 100 & 25.72 & 24.52 & 23.84 & 23.43 & 23.28 \\ 
\textit{CW $\ell_\infty$} & 100 & 22.8  & 21.39 & 21.21 & 20.81 & 20.92 \\  \hline
\end{tabular}
}
\caption{Classification error (\%) on the first 1000 test samples for the DCT domain with the local sign flipping in 3 sub-bands {\small($J = 3$)}.}
\label{tab:dct_sign_classification_error}
\end{table}
%
%-------------------------------

The situation with the MNIST data set is even more interesting. First of all, we would like to point out that we decrease in 2 times the classification error in comparison to the classical scenario. However, for the \textit{CW $\ell_0$} the results are surprisingly worse. To investigate the reasons of the performed degradation we visualize the adversarial examples. The results are shown in Table \ref{tab:adversarial_noise}. It is easy to see that, in general, the \textit{CW $\ell_\infty$} noise manifests itself as a background distortion and doesn't affect considerably the regions of useful information. The \textit{CW $\ell_2$} noise affects the regions of interest but the intensity of the noise is much lower than the intensity of the meaningful information. As well as in \textit{CW $\ell_2$}, the \textit{CW $\ell_0$} noise is concentrated in the region near the edges but its intensity is as strong as the informative image parts. Thus, it becomes evident why local DCT sign permutation is not capable to withstand such kind of noise. In general, such a strong noise is easy detectable and the corresponding adversarial examples can be rejected by many detection mechanisms, like for example an auxiliary "detector" sub-network \cite{metzen2017detecting}. Moreover, as it can be seen from the Fashion-MNIST examples, the influence of such noise and successful attacks drastically decreases with increasing image complexity. As it has been shown by the CIFAR-10 results, the local DCT sign permutation produces a high level of defense against such an attack for natural images.

%-----------------------------------------
%
\begin{table}[t!]
\renewcommand*{\arraystretch}{1.25}
\begin{center}
\begin{tabular}{ccc} \hline
Attack & MNIST & Fashion-MNIST \\ \hline
& & \\[-0.45cm] 
\multirow{ 2}{*}{\textit{CW $\ell_2$}} & \begin{minipage}{0.075\textwidth}\includegraphics[width=0.75\textwidth]{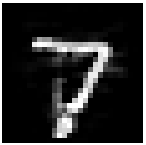}\end{minipage} & 
\begin{minipage}{0.075\textwidth}\includegraphics[width=0.75\textwidth]{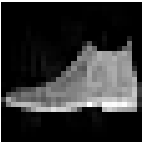}\end{minipage} \\
& \begin{minipage}{0.075\textwidth}\includegraphics[width=0.75\textwidth]{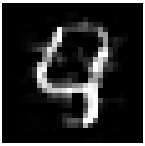}\end{minipage} & 
\begin{minipage}{0.075\textwidth}\includegraphics[width=0.75\textwidth]{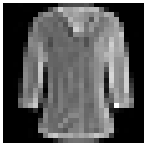}\end{minipage} \\
& & \\[-0.45cm] \hline
& & \\[-0.45cm] 
\multirow{ 2}{*}{\textit{CW $\ell_0$}} & \begin{minipage}{0.075\textwidth}\includegraphics[width=0.75\textwidth]{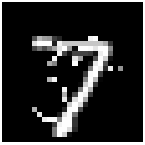}\end{minipage} 
& \begin{minipage}{0.075\textwidth}\includegraphics[width=0.75\textwidth]{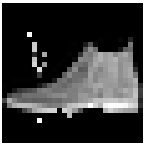}\end{minipage} \\
& \begin{minipage}{0.075\textwidth}\includegraphics[width=0.75\textwidth]{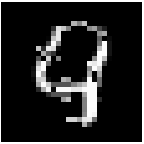}\end{minipage}  
& \begin{minipage}{0.075\textwidth}\includegraphics[width=0.75\textwidth]{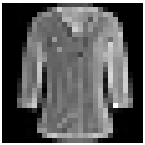}\end{minipage}\\ 
& & \\[-0.45cm] \hline
& & \\[-0.45cm] 
\multirow{ 2}{*}{\textit{CW $\ell_\infty$}} & \begin{minipage}{0.075\textwidth}\includegraphics[width=0.75\textwidth]{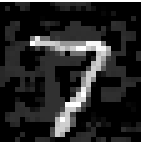}\end{minipage} 
& \begin{minipage}{0.075\textwidth}\includegraphics[width=0.75\textwidth]{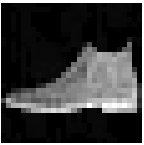}\end{minipage} \\ 
& \begin{minipage}{0.075\textwidth}\includegraphics[width=0.75\textwidth]{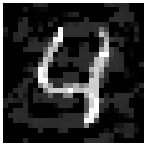}\end{minipage} 
& \begin{minipage}{0.075\textwidth}\includegraphics[width=0.75\textwidth]{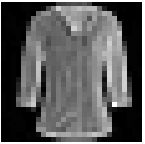}\end{minipage} \\ 
& & \\[-0.45cm] \hline
\end{tabular}
\end{center}
\caption{Adversarial examples.}
\label{tab:adversarial_noise}
\end{table}
%
%-----------------------------------------
%

%
%-------------------------------------------------------------------------
\section{Classification with multi-channel hard thresholding in the sub-bands of the DCT domain}
\label{sec:zero filling as a defense strategy}

As it can be seen from Figure \ref{fig:dct_sign_permut}, the local DCT sign permutation creates sufficiently high image distortions. As a simple strategy to avoid this effect, we investigate hard thresholding of the DCT coefficients in the defined sub-bands. In this case the matrix $\mathbb P_{ji}$ contains zeros for the coefficients of key-defined sub-bands. Alternatively, one can consider this strategy as a random sampling as illustrated in Figure \ref{fig:p_sampling}, where one retains only the coefficients used by the classifier. In this sense, the considered strategy is close to the randomization in a single channel without the aggregation considered in \cite{chen2018secure}.

Note that the considered processing  is a data independent transform. The secret keys can be used for choosing the sub-bands positions. Thus, the attacker can not predict in advance, which DCT coefficients will be used or suppressed.

\begin{figure}[t!]
     \centering
     \subfloat[][original]{\quad\includegraphics[width=0.125\linewidth]{images/006_3_original.png}\label{fig:permut_dct_original}\quad}  
     \hspace{0.2cm}
     \subfloat[][sub-band V]{\quad\includegraphics[width=0.125\linewidth]{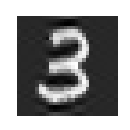}\label{fig:permut_dct_v_zeros}\quad}
     \hspace{0.2cm}
     \subfloat[][sub-band H]{\quad\includegraphics[width=0.125\linewidth]{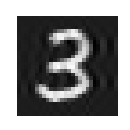}\label{fig:permut_dct_h_zeros}\quad}  
     \hspace{0.2cm}
     \subfloat[][sub-band D]{\quad\includegraphics[width=0.125\linewidth]{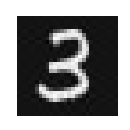}\label{fig:permut_dct_d_zeros}\quad}         
     \caption{Local zero filling in the DCT domain.}
     \label{fig:dct_zero_filling}
\end{figure}

For simplicity and to be comparable with the previously obtained results, we use the multi-channel architecture that is shown in Figure \ref{fig:classification_via_dct_permut},  the DCT sub-band division as illustrated in Figure \ref{fig:dct_subbands} with fixed 3 sub-band sizes and positions. Instead of applying the sign permutation, the corresponding DCT frequencies are set to zero and the result is transformed back to the direct domain. The visualization of the results of such a transformation is shown in Figure \ref{fig:dct_zero_filling}. The resulting images are slightly blurry but less noisy than in the case of the DCT sign permutation. 

The obtained numerical results for the MNIST, Fashion-MNIST and CIFAR-10 data sets are given in Table \ref{tab:dct_zero_filling}. In general, the results are very close to the results of using the DCT sign permutation represented in Table \ref{tab:dct_sign_classification_error} with the number of classifiers equals to 3. For the original non-attacked data the classification error is almost the same. In case of the attacked data, the  classification error is about 0.5-1 \% higher. This is related to the fact that the zero replacement of DCT coefficients leads to a loss of information and, consequently, to a decrease in classification accuracy. 

Hence, replacing the DCT coefficients by zeros might also serve as a defense strategy. 

%-----------------------------------------
%
%

%-------------------------------------------------------------------------
\section{Conclusions}
\label{sec:conclusions}

In this paper, we address a problem of protection against adversarial attacks in classification systems.  
%We first briefly discussed the state-of-the-art defense strategies. 
We propose the randomized diversification mechanism as a defense strategy in the multi-channel architecture with the aggregation of classifiers' scores. The randomized diversification is a secret key-based randomization in a defined domain. The goal of this randomization is to prevent the gradient back propagation or use of bypass systems by the attacker. We evaluate the efficiency of the proposed defense and the performance of several variations of a new architecture on three standard data sets against a number of known state-of-the-art attacks. The numerical results demonstrate the robustness of the proposed defense mechanism against adversarial attacks and show that using the multi-channel architecture with the following aggregation stabilizes the results and increases the classification accuracy. 

For the future work we aim at investigating  the proposed defense strategy against the gradient based sparse attacks and non-gradient based attacks.

\begin{table}[t!]
{\small
\renewcommand*{\arraystretch}{1.25}
\begin{center}
\begin{tabular}{lcccc} \hline
& Original & \textit{CW $\ell_2$} & \textit{CW $\ell_0$} & \textit{CW $\ell_\infty$} \\ \hline
MNIST         & 0.6  & 7.59  & 21.3 & 3.03 \\
Fashion-MNIST & 8.8  & 9.6   & 11.23 & 9.58 \\
CIFAR         & 21.1 & 23.28 & 27.08 & 23.27 \\ \hline
\end{tabular}
\end{center}
}
\caption{Classification error (\%) for the DCT based hard thresholding over the first 1000 test samples {\small($J = 3, \; I = 1$)}.}
\label{tab:dct_zero_filling}
\end{table}
%
%-----------------------------------------
%

%\clearpage
%-------------------------------------------------------------------------
%\newpage

{\small
\bibliographystyle{ieee}
\bibliography{cvpr_2019}
}

\end{document}